\documentclass{article}

\usepackage{arxiv}

\usepackage[utf8]{inputenc} 
\usepackage[T1]{fontenc}    
\usepackage{hyperref}       
\usepackage{url}            
\usepackage{booktabs}       
\usepackage{amsfonts}       
\usepackage{nicefrac}       
\usepackage{microtype}      
\usepackage{lipsum}
\usepackage{graphicx}
\graphicspath{ {./images/} }

\usepackage{float}
\usepackage{cite} 
\usepackage{float}
\usepackage{graphicx}
\usepackage{bbding} 
\usepackage{colortbl}
\usepackage{CJKutf8}
\usepackage{makecell}
\usepackage[figuresright]{rotating}
\usepackage{booktabs}
\usepackage{wrapfig}
\usepackage{times}
\usepackage{latexsym}
\usepackage[T1]{fontenc}
\usepackage[utf8]{inputenc}
\usepackage{graphicx} 
\usepackage{subfigure}
\usepackage{microtype}
\usepackage{amsmath,amsfonts}
\usepackage{algorithmic}
\usepackage{array}
\usepackage{textcomp}
\usepackage{stfloats}
\usepackage{url}
\usepackage{verbatim}
\usepackage{graphicx}
\usepackage{balance}
\usepackage{multirow}
\usepackage{float}
\usepackage{cite} 
\usepackage{float}
\usepackage{graphicx}
\usepackage{bbding} 
\usepackage{colortbl}
\usepackage{CJKutf8}
\usepackage{makecell}
\usepackage[figuresright]{rotating}
\usepackage{booktabs}
\usepackage{wrapfig}

\usepackage[utf8]{inputenc} 
\usepackage[T1]{fontenc}    
\usepackage{hyperref}       
\usepackage{url}            
\usepackage{booktabs}       
\usepackage{amsfonts}       
\usepackage{nicefrac}       
\usepackage{microtype}      
\usepackage{xcolor}         

\usepackage[utf8]{inputenc} %
\usepackage[T1]{fontenc}    %
\usepackage{hyperref}       %
\usepackage{url}            %
\usepackage{booktabs}       %
\usepackage{amsfonts}       %
\usepackage{nicefrac}       %
\usepackage{microtype}      %
\usepackage{xcolor}         %
\usepackage{lipsum}
\usepackage{graphicx}
\usepackage{wrapfig}
\usepackage[size=small]{caption}
\usepackage{mathtools}
\usepackage{amssymb}
\usepackage{amsthm}
\usepackage{wrapfig}
\usepackage[algoruled,boxed,lined,noend]{algorithm2e}
\usepackage{cancel}
\usepackage{listings}
\usepackage{color}

\definecolor{dkgreen}{rgb}{0,0.6,0}
\definecolor{gray}{rgb}{0.5,0.5,0.5}
\definecolor{mauve}{rgb}{0.58,0,0.82}

\title{COPR: Continual Learning Human Preference through Optimal Policy Regularization}

\author{Han Zhang$^{1,2}$, Lin Gui$^{3}$, Yuanzhao Zhai$^{4,2}$\\
\textbf{Hui Wang}$^{2}$, \textbf{Yu Lei}$^{2*}$, \textbf{Ruifeng Xu}$^{1,2}$\thanks{\quad Corresponding author.} \\
  $^{1}$Harbin Institute of Technology (Shenzhen), Shenzhen 518000, China \\
  $^{2}$Peng Cheng Laboratory, Shenzhen 518000, China \\
  $^{3}$King’s College London, London WC2R 2LS, United Kingdom \\
  $^{4}$National University of Defense Technology, Changsha 410000, China \\
  \texttt{hanlardresearch@gmail.com, lin.1.gui@kcl.ac.uk, yuanzhaozhai@126.com} \\
  \texttt{\{wangh06, leiy01\}@pcl.ac.cn, xuruifeng@hit.edu.cn}
}

\begin{document}
\maketitle

\begin{abstract}
The technique of Reinforcement Learning from Human Feedback (RLHF) is a commonly employed method to improve pre-trained Language Models (LM), enhancing their ability to conform to human preferences. Nevertheless, the current RLHF-based LMs necessitate full retraining each time novel queries or feedback are introduced, which becomes a challenging task because human preferences can vary between different domains or tasks. Retraining LMs poses practical difficulties in many real-world situations due to the significant time and computational resources required, along with concerns related to data privacy. 
To address this limitation, we propose a new method called Continual Optimal Policy Regularization (COPR), in which we compute the distribution of optimal policy bypassing the partition function and then regularize the current policy based on the historically optimal distribution to mitigate Catastrophic Forgetting (CF).  
COPR involves a single learning phase and doesn't necessitate complex reinforcement learning. 
Importantly, it shares the capability with RLHF to learn from unlabeled data by maintaining a scoring module, similar to reward model, making it flexible for continually learning without human feedback. 
Our experimental results show that COPR outperforms strong Continuous Learning (CL) baselines when it comes to consistently aligning with human preferences on incremental tasks and domains.
\end{abstract}

\section{Introduction}
In the realm of natural language processing (NLP), large language models (LLMs) are vital tools with the potential to bridge human language and machine understanding. Learning human preferences is a crucial step towards ensuring that language models not only generate responses that are useful to users but also adhere to ethical and societal norms, namely helpful and harmless responses \cite{HH_RLHF}. However, they face a fundamental challenge in aligning with human preferences and values, hindering their full potential.

Traditional alignment methods, namely Reinforcement Learning from Human Feedback (RLHF)\cite{summary_rlhf,InstructGPT}, involve supervised fine-tuning (SFT), reward model (RM) training, and policy model training. 
This complex pipeline lacks flexibility for continual learning (CL) of human preferences, hence existing work\cite{HH_RLHF} often necessitates retraining models to adapt to dynamic preferences, which needs huge computation resources. 
Hence, there is a pressing need for research into continual alignment methods to address this limitation, enabling LLMs to better adhere to evolving human preferences and values while generating helpful responses.

\begin{figure}[h]
    \centering
    \subfigure[framework]{
     \label{fig:framework}
    \includegraphics[width=0.45\linewidth, trim=125 300 495 0,clip]{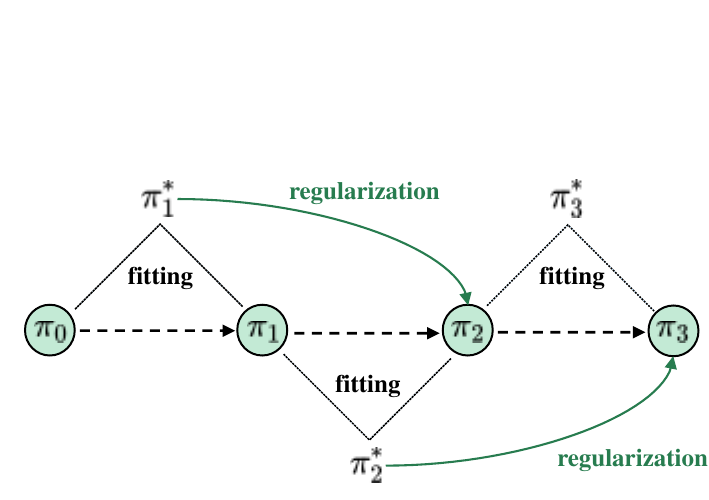}
    }
    \subfigure[taxonomy]{
    \label{fig: taxonomy}
     \includegraphics[width=0.45\linewidth, trim=0 320 500 0,clip]{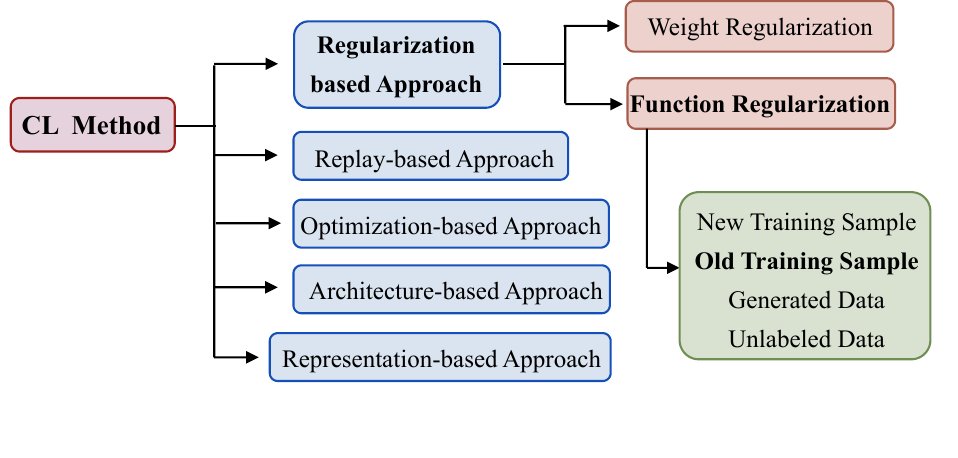}
    }
    \caption{\textbf{(a)} The framework of COPR. The optimal policy $\pi_t^{*}$ $(t=1,2,3)$ is derived from the policy $\pi_{t-1}$. The optimal policy $\pi_t^{*}$ is utilized as the fitting objective of $\pi_{t}$ and the regularization term of $\pi_{t+1}$. (b) A state-of-the-art and elaborated taxonomy \cite{SurveyCL2023} of representative continual learning methods. Bold indicates the category to which our method belongs.}
   
\end{figure}
In this paper, we propose an innovative approach to \textbf{C}ontinually learning human preference through \textbf{O}ptimizal \textbf{P}olicy \textbf{R}egularization (COPR).
Our method, as depicted in Figure  \ref{fig:framework}, computes the sequence of optimal policy  distributions in scenarios where human preferences evolve continuously. 
Then, we use the optimal policy distributions as supervision signal and regularization objective to finetune the policy model. 
Intuitively, the COPR makes the distribution of the policy close to the optimal policy on the current task and prevents it from deviating too far from the historically optimal policies for past tasks. 
According to the state-of-the-art continual learning taxonomy\cite{SurveyCL2023}, our method belongs to the category of  \textit{function regularization}. 
Inspired by the reward model in the RLHF pipeline, we maintain a scoring module to give the preference order for responses, which can learn unlabeled data.

To the best of our knowledge, we are the first to study the CL of alignment methods. For fair evaluation, 
we construct the first Task Incremental Learning (TIL) and Domain Incremental Learning (DIL) benchmarks for continual learning of human preferences based on existing human preference datasets (Section \ref{sec:benchmark}). 
The TIL benchmark utilizes Helpful and Harmless preference (HH)\cite{HH_RLHF} data, Reddit TL;DR summary\cite{summary_rlhf} human preference data provided by CarperAI \footnote{ 
For each Reddit post in the dataset, multiple summaries are generated using various models. These models include pre-trained ones used as zero-shot summary generators, as well as supervised fine-tuned models (12B, 6B, and 1.3B) specifically trained on the Reddit TL;DR dataset. Additionally, the human-written TL;DR (reference) is considered as a sample for comparison.
\textbf{URL}: \url{https://huggingface.co/datasets/CarperAI/openai_summarize_comparisons} },  the \textit{enhanced} IMDB\cite{IMDB} Sentiment Text Generation benchmark released by RL4LMs \cite{NLPO}.
The DIL benchmark utilizes the Standard Human Preference (SHP)\cite{shp} data which has 18 domains including academia, baking, physics, etc.

In summary, our main contribution is: 
\begin{itemize}
    \item We propose COPR, a simple RL-free algorithm for continually learning human preferences. 
    \item We introduce the first TIL and DIL benchmarks for continual value alignment.
    \item Our experiments show that COPR outperforms existing CL and alignment methods on TIL and DIL benchmarks.
\end{itemize}

\section{Preliminaries}
\subsection{Static Alignment}
\textbf{Reinforcement Learning from Human Feedback}.
The recent RLHF pipeline consists of three phases: 1) supervised fine-tuning (SFT); 2) preference sampling and reward learning and 3) reinforcement-learning optimization.
In the SFT phases, the language model is fine-tuned with supervised learning (maximum likelihood) on the downstream tasks.
In the reward learning phase, human annotators rank multiple answers $\{y_1^x \prec y_2^x \prec...\prec y_{n}^x\}$ for a prompt x based on human preferences, generating human feedback data. Then, this feedback data is used to train a reward model, which assigns higher scores to pairs consisting of prompts and answers that are preferred by humans.
In the RL Fine-Tuning phase,  the mainstream methods maximize a KL-constrained reward objective like
\begin{align}\label{eq:RL}
\max_{\pi_{\theta}}  \mathbb{E}_{x\sim \mathcal{D}, y\sim \pi_{\theta}(y \mid x)}\bigl[r_{\phi}(x, y)\bigr] - \beta\mathbb{D}_{\textrm{KL}}\bigl[\pi_{\theta}(y\mid x)\mid \mid 
\pi_{ref}(y\mid x)\bigr]
\end{align}
where $\beta$ is a parameter controlling the deviation from the base reference policy $\pi_{ref}$, namely the initial SFT model $\pi_{sft}$. 
Because language generation operates discretely, this objective lacks differentiability and is generally optimized using reinforcement learning techniques. 
The recent approaches \cite{HH_RLHF,InstructGPT,summary_rlhf}  reconstruct the reward function ${r(x, y) = r_{\phi}(x, y) -\beta (\log \pi_{\theta}(y\mid x) - \log \pi_{ref}(y\mid x))}$, and maximize using PPO \cite{PPO}. 

\textbf{The Theory of Optimal Policy}. 
Previous work DPO\cite{DPO} proposes that the optimal solution to the KL-constrained reward maximization objective in Eq. \ref{eq:RL} takes the form:
\begin{equation}\label{eq:optimsol}
    \pi_r(y\mid x) = \frac{1}{Z(x)}\pi_{ref}(y\mid x)\exp\left(\frac{1}{\beta}r(x, y)\right)
\end{equation}
where $Z(x) =\sum_{y}\pi_{ref}(y\mid x)\exp\left(\frac{1}{\beta}r(x, y)\right)$ is the partition function. 

\subsection{Continual Alignment}
In the static alignment, the dataset typically consists of a fixed, static set of examples that are collected and labeled for a specific task. The dataset remains constant throughout the training process.
In the continual alignment scenario, the human preference dataset evolves over time, often consisting of a sequence of tasks or domains. Each task or domain may have its own set of data, and these tasks are presented to the model sequentially. The order in which tasks are presented can vary, and the model needs to adapt to new tasks without forgetting previously learned ones.

In this paper, we consider that there is a sequence of tasks $\mathbb{T} = \{\mathcal{T}_1, \mathcal{T}_2, ... \}$ to learn, and a sequence of corresponding human preference datasets  $\mathbb{D} =\{\mathcal{D}_1,\mathcal{D}_2,...\}$. The initial policy is the SFT model, namely, $\pi_0 = \pi_{SFT}$. For each task $\mathcal{T}_t (t=1,2,...)$,  the policy  $\pi_{t}$ is initialized by $\pi_{t-1}$ and there is a latent scoring function (i.e. the preference model) $r_t(x,y)$.  For each prompt $x$,  the partially-ordered set of  responses  $\mathcal{Y}^x = \{y_1^x \prec y_2^x \prec...\prec y_{J_x}^x \}$ is known.  If learning with human feedback, the order represents the degree of human preference; otherwise, the order information is obtained from the ranking scores assigned by the preference model. 
To mitigate forgetting, a replay memory buffer $\mathbb{R} =\{\mathcal{R}_1,\mathcal{R}_2,...\}$ is maintained, where $\mathcal{R}_i \subset \mathcal{D}_i$ ($i=1,2,...,t$)  stores training data from 1\% of historical tasks. When learning new tasks, the data in the replay memory is merged with the training data of the new task.

\section{Method}

\subsection{Continual Learning by Optimal Policy Regularization}
\label{method}
Previous works\cite{DPO,AWR}, prove that the optimal solution to the KL-constrained reward maximization objective takes the form of Eq. \ref{eq:optimsol}. Based on this, we conclude that the optimal policy of task $\mathcal{T}_t$ is 
\begin{equation}
    \pi^{*}_{t}(y|x)=\frac{1}{Z_t(x)}\pi_{t-1}(y|x)exp(\frac{1}{\beta}r_t(x,y))
    \label{eq:optimPI}
\end{equation}
where $Z_t(x)=\Sigma_{y}\pi_{t-1}(y|x)exp(\frac{1}{\beta}r_t(x,y))$ is the partition function, $x \in \mathcal{D}_t$ denotes the prompt, $y\in$ denotes the possible response. For the estimation $\pi^{*}_{t}(y|x)$, we suppose that there are $J_x$ responses  of each prompt $x$ and the partial order annotated by humans is $y_1^x \prec y_2^x \prec...\prec y_{J_x}^x$. 
The COPR method includes 4 steps, construct reward function, calculate the sampling distribution, optimal distribution fitting, and optimal distribution regularization. 


 \textbf{Step 1: Construct reward function}

 We split the reward $r(x,y)$ into the expected reward $\delta(x)$ and the advantage score $Adv(x,y)$, i.e., the extra reward one response can obtain compared with the expected reward.
\begin{equation}
    r_t(x,y_j^x) = Adv(x,y_j^x) + \delta(x)
    \label{eq:reward}
\end{equation}
for approximating the well trained reward function, where $j=1,2,...,{J_x}$ represents the degree of human preference, the expectation $\delta(x)=E_{y\sim\pi({\cdot|x})} r(x,y)$ depends solely on the prompt $x$. We attempt to model the advantage score by a \textit{Linear} function or a \textit{Gaussian} distribution and regulate the advantage score distribution dynamically during training, ensuring that the variances and means are maintained within a reasonable range. 

\begin{itemize}
    \item \textbf{Linear reward}: By deriving the gradient of the pairwise loss function $\mathcal{L}_{ranking}=-\log(\sigma(r_\theta(x,y_w)-r_\theta(x,y_l)))$, we found that the reward scores are approximately linearly related to human preferences.\footnote{As the training steps increase, the reward scale rises and the score function tends to a non-linear style. Because the sigmoid function has an approximate linear region. When the reward values are within this region, the gradients also approximately linearly increase with the degree of preference. When the reward values reach the saturation region of the sigmoid, the gradients no longer increase approximately linearly with the degree of preference.}  We provide theoretical derivations in the Appendix Section \ref{sec:pairwise}.Previous works \cite{LinearRM2020,LinearRM2017,LinearRM2012,LinearRM2014} present the reward as a linear combination of pre-trained features or hand-crafted features.  Recent work\cite{linearRM} gets enhanced performance by modeling a linear reward function according to \textit{regret} which is the negated sum of an optimal policy’s advantage in the segment.  Inspired by this,  we propose to use a linear advantage function  $Adv(x,y_j^x)\triangleq\frac{2j-J_x-1}{J_x} \in (-1,1)$. Linear advantage assigns scores to each response in an arithmetic progression, effectively translating the human preference order into rewards. 
\end{itemize}

\begin{itemize}
    \item \textbf{Gaussian reward}: The Gaussian distribution is also observed in well-trained reward scores\cite{peng2023stabilizing}. Inspired by this observation, we attempt to model the advantage score by a normal distribution. We sample $J_x$ values $\{R_1 < R_2 < ... < R_{J_x}  \}$ from the distribution $N(0,\sigma)$, and define $Adv(x,y_j^x)\triangleq{R_j}$. In Step 2, we theoretically prove that there is no need to calculate specific values for $\delta(x)$ and partition function  $Z(x)$.
\end{itemize}

 \textbf{Step 2: Calculate the Sampling Distribution  }$P^{*}_{y \in \mathcal{Y}^x, t}(y|x)$.  
 
Given the partially-ordered set of responses $\mathcal{Y}^x = \{y_1^x \prec y_2^x \prec...\prec y_{J_x}^x \}$. We calculate the re-normalized distribution $P^{*}_{y \in \mathcal{Y}^x,t}(y|x)$:
\begin{equation}
\begin{aligned}
P^{*}_{y \in \mathcal{Y}^x,t}(y|x) &\triangleq  \frac{\pi^{*}_{t}(y|x)}{ \Sigma_{y^{'} \in \mathcal{Y}^x}\pi^{*}_{t}(y^{'}|x)} \\
&= \frac{\cancel{\frac{1}{Z_t(x)}} \cdot \pi_{t-1}(y|x)\cdot exp(\frac{1}{\beta} Adv(x,y))\cdot \cancel{exp(\frac{1}{\beta}\delta(x))}}{\Sigma_{y^{'} \in \mathcal{Y}^x} \cancel{\frac{1}{Z_t(x)}}\cdot \pi_{t-1}(y^{'}|x) \cdot exp(\frac{1}{\beta}Adv(x,y^{'})) \cdot \cancel{exp(\frac{1}{\beta}\delta(x))} } \\
&= \frac{\pi_{t-1}(y|x)exp(\frac{1}{\beta}Adv(x,y))}{\Sigma_{y^{'} \in \mathcal{Y}^x}\pi_{t-1}(y^{'}|x) exp(\frac{1}{\beta}Adv(x,y^{'})) }
\end{aligned}
\end{equation}
Please note that the partition function $Z(x)$ is canceled out during the calculation, which simplifies the optimization objective.  Because evaluating $Z(x)$ \cite{Fdivergence} is challenging and often introduces new errors\cite{goodfellow2016deep}. 
Furthermore, the sampling distribution is also independent of reward expectation $\delta(x)$, which means COPR is \textit{invariant} with respect to the \textit{equivalent reward}\cite{P3O}.

\textbf{Step 3: Fit the distribution  $P^{*}_{y \in \mathcal{Y}^x,t}(y|x)$}.

Next, we directly fit the re-normalized probability $P^{*}_{y \in \mathcal{Y}^x,t}(y|x)$ by minimizing the KL loss task $\mathcal{T}_t$:
\begin{equation}
    \begin{aligned}
        L_{t}^{fit}({\theta}_t) &= \mathbb{E}_{x\sim {\mathcal{D}_t}}\Sigma_{y \in \mathcal{Y}^x } D_{KL}(P_{y \in \mathcal{Y}^x,t}(y|x,{\theta}_t), P^{*}_{y \in \mathcal{Y}^x,t}(y|x) )
    \end{aligned}
\end{equation}

where $\theta_t$ denotes the parameters of the policy model $\pi_{t}(y|x)$ at task  $\mathcal{T}_t$ $(t=1,2,...)$, and $P_{y \in \mathcal{Y}^x,t}(y|x,{\theta})$  denotes the re-normalized probability of current model:
\begin{align}
P_{y \in \mathcal{Y}^x,t}(y|x,{\theta}_t) \triangleq \frac{\pi_{t}(y|x)}{ \Sigma_{y^{'} \in \mathcal{Y}^x}\pi_{t}(y^{'}|x)} 
\end{align}
\begin{wrapfigure}[3]{r}{5cm}
    \centering
    \includegraphics[width=5cm]{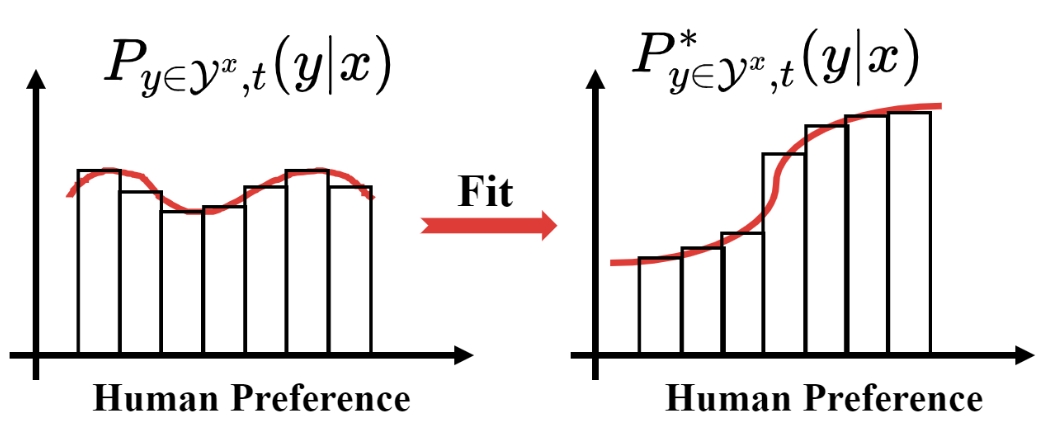}
    \caption{Fitting of  $P^{*}_{y \in \mathcal{Y}^x,t}(y|x)$}
    \label{fig:fitP}
\end{wrapfigure}
Steps 2-3 can be implemented by 4 lines of codes under the PyTorch environment:
\begin{lstlisting}
scale_opt = (1 / beta * rm_score.detach()).exp()
P_optimal = P_ref * scale_opt / (P_ref * scale_opt).sum()
P_policy = P_current / P_current.sum()
RDPO_loss = F.kl_div(P_policy.log(), P_optimal)
\end{lstlisting}

\textbf{Step-4: Function Regularization of $\pi_t$}.

To preserve the old knowledge, we calculate a regularization loss to ensure that the new and old optimal policies do not differ too significantly in terms of the distribution of old human preferences.
In detail, for each replay sample $x \in {\mathcal{R}_i}$ $(i=1,2,...,t-1)$ , the new policy $\pi_t$ is regularized to not differ significantly from the optimal policy $\pi_i^{*}$. Hence, the regularization loss is
\begin{equation}
    L_{t}^{reg}({\theta}_t) = \mathbb{E}_{x\sim {\cup_{i=1}^{i=t-1}\mathcal{R}_i}} \Sigma_{i=1}^{t-1} \boldsymbol{I_{\mathcal{R}_i}}(x) \cdot\Sigma_{y \in \mathcal{Y}^x } D_{KL}(P_{y \in \mathcal{Y}^x,t}(y|x,{\theta}_t),  P^{*}_{y \in \mathcal{Y}^x,i}(y|x) )
\end{equation}
where $\{\boldsymbol{I_{\mathcal{R}_i}}(x)\}_{i=1}^{i=t-1}$ is the set of indicator functions, namely, the task identifier.
The final training loss of task $\mathcal{T}_t$ is: 
\begin{equation}
    L_{t}^{train}({\theta}_t) = 
\left\{
     \begin{array}{lr}
     L_{t}^{fit}({\theta}_t) &  x \in {\mathcal{D}_t}    \\
     L_{t}^{reg}({\theta}_t) &  x \in {\cup_{i=1}^{i=t-1}\mathcal{R}_i}  
     \end{array}
\right.
\end{equation}

\subsection{Incremental Learning on Unlabeled Data}
In the above steps of COPR, we learn a policy based on the static human preference dataset without additional evaluations or generations, namely, offline learning. 
Inspired by the reward model in RLHF, we add a preference scoring module to offline COPR, to score and rank the responses instead of humans. 
We find that the preference scoring module is compatible with the policy model, and does not need to learn an independent preference model, like the RLHF pipeline. 
In detail, we introduce a value head $\mathcal{V}$ and utilize the pairwise ranking loss $\mathcal{L}_{ranking}$ to learn an RM score, where $\mathcal{L}_{ranking}=-\log(\sigma(r_\theta(x,y_w)-r_\theta(x,y_l)))$, where $y_w$ and $y_l$ represent the human choose and human rejected response respectively.  
The final training loss is the sum of  $L_{t}^{train}({\theta}_t)$ and $\mathcal{L}_{ranking}$.
After learning the labeled data, the reward value head can be used to score the unlabeled data and rank unlabeled responses.  Based on this ranking, we can utilize the COPR method. 
Due to Collecting human preference data being expensive, utilizing a model to score instead of humans has great practical application prospects in real-world scenarios. 
In Section \ref{sec:unlabel}, we experimentally evaluate this method on the SHP\cite{shp} dataset.

\subsection{Comparison with other methods}

We compare COPR with recent alignment methods in table \ref{tab:compare}.
Our approach employs the same optimal policy form as DPO, the main difference is that DPO uses the log ratio as a reward value, while COPR uses a custom reward (linear or gaussian), and DPO maximizes the gap between wins and losses, while COPR fits the distribution of the optimal policy on the sampled dataset.  
PRO also computes the sampling distribution in their objective, the main difference with COPR is that PRO enhances the probability of top-ranked samples occupying all sampled instances, while COPR fits the optimal policy distribution.  
\begin{table}[h]
\caption{Compare COPR with other alignment methods.}
\resizebox{\linewidth}{!}{
\begin{tabular}{ccccccccccc}
\toprule
\multicolumn{1}{l}{\textbf{Method}} & \multicolumn{1}{l}{\textbf{\begin{tabular}[c]{@{}l@{}}RL or \\ Non-RL\end{tabular}}} & \multicolumn{1}{l}{\textbf{\begin{tabular}[c]{@{}l@{}}Online or \\ Offline\end{tabular}}} & \multicolumn{1}{l}{\textbf{\begin{tabular}[c]{@{}l@{}}Pairwise or\\ Listwise\end{tabular}}} & \multicolumn{1}{l}{\textbf{\begin{tabular}[c]{@{}l@{}}Token-wise or \\ Trajectory-wise\end{tabular}}} & \multicolumn{1}{l}{\textbf{\begin{tabular}[c]{@{}l@{}}Invariance \\ \cite{P3O} \end{tabular}}} & \multicolumn{1}{l}{\textbf{\begin{tabular}[c]{@{}l@{}}Reward \\ Model\end{tabular}}} & \multicolumn{1}{l}{\textbf{\begin{tabular}[c]{@{}l@{}}Reference \\ Model\end{tabular}}} & \multicolumn{1}{l}{\textbf{\begin{tabular}[c]{@{}l@{}}Critic \\ Model\end{tabular}}} & \multicolumn{1}{l}{\textbf{\begin{tabular}[c]{@{}l@{}}Num of \\ Models\end{tabular}}} & \multicolumn{1}{l}{\textbf{\begin{tabular}[c]{@{}l@{}}Continual \\ Learning\end{tabular}}} \\ \midrule
\textbf{PPO\cite{PPO}}              & RL                                                                                   & online                                                                                    & pairwise                                                                                   & token-wise                                                                                            & no                                                                                           & yes                                                                                  & yes                                                                                      & yes                                                                                  & 4                                                                                     & no                                                                                         \\
\textbf{NLPO\cite{NLPO}}            & RL                                                                                   & online                                                                                    & pairwise                                                                                   & token-wise                                                                                            & no                                                                                           & yes                                                                                  & yes                                                                                      & yes                                                                                  & 4                                                                                     & no                                                                                         \\
\textbf{P3O\cite{P3O}}              & RL                                                                                   & online                                                                                    & pairwise                                                                                    & trajectory-wise                                                                                       & yes                                                                                          & yes                                                                                  & yes                                                                                      & no                                                                                   & 3                                                                                     & no                                                                                         \\
\textbf{PRO\cite{PRO}}             & non-RL                                                                               & offline                                                                                   & listwise                                                                                    & trajectory-wise                                                                                       & -                                                                                            & no                                                                                   & no                                                                                       & no                                                                                   & 1                                                                                     & no                                                                                         \\
\textbf{DPO\cite{DPO}}              & non-RL                                                                               & offline                                                                                   & pairwise                                                                                    & trajectory-wise                                                                                       & -                                                                                            & no                                                                                   & yes                                                                                      & no                                                                                   & 2                                                                                     & no                                                                                         \\
\textbf{RAFT\cite{raft}}           & non-RL                                                                               & both                                                                                   & listwise                                                                                    & trajectory-wise                                                                                       & yes                                                                                          & yes                                                                                  & no                                                                                       & no                                                                                   & 2                                                                                     & no                                                                                         \\
\textbf{RRHF\cite{rrhf}}         & non-RL                                                                               & offline                                                                                   & listwise                                                                                    & trajectory-wise                                                                                       & yes                                                                                          & yes                                                                                  & no                                                                                       & no                                                                                   & 2                                                                                     & no                                                                                         \\ 
\textbf{CoH\cite{CoH}}	 & non-RL                                                                               & offline                                                                                   & pairwise                                                                                    & trajectory-wise                                                                                       & -                                                                                            & no                                                                                   & no                                                                                       & no                                                                                   & 1                                                                                     & no                                                                                         \\
\midrule
\textbf{COPR}                       & non-RL                                                                               & both                                                                                   & listwise                                                                                    & trajectory-wise                                                                                       & -                                                                                            & no                                                                                   & yes                                                                                      & no                                                                                   & 2                                                                                     & yes                                                                                        \\ \bottomrule
\end{tabular}}
\label{tab:compare}
\end{table}

\section{Experiments}

\subsection{Benchmarks, Tasks and Evaluation Metrics}
\label{sec:benchmark}

We build the first TIL and DIL benchmarks based on existing human preference datasets including the HH-RLHF\cite{HH_RLHF} dataset, Reddit TL;DR summary\cite{summary_rlhf} human preference data, the IMDB\cite{IMDB} Sentiment Text Generation dataset \cite{NLPO}, and the SHP\cite{shp} dataset, which are detailed in Appendix Section \ref{sec:dataset}.

\textbf{Task Incremental Learning (TIL)} setting:  The policy is required with continually learning across three distinct tasks: the QA task on the HH-RLHF\cite{HH_RLHF} dataset, the summary task on the Reddit TL;DR human feedback \cite{TLDR} dataset and the positive file review generation task on the IMDB\cite{IMDB} dataset.
As shown in Table \ref{tab:metrics4TIL}, we utilize 3 preference metrics and 3 naturalness metrics to evaluate the performance of the model. In detail, we employ the \textit{SteamSHP-flan-t5-xl model} \cite{shp}, developed by Stanford, as the preference model (PM) for assessing responses to HH-RLHF prompts. Additionally, we utilize the 6.7B \textit{gpt-j} reward model \footnote{URL: \url{https://huggingface.co/CarperAI/openai_summarize_tldr_rm_checkpoint}}, released by Carper-AI, to evaluate summaries of Reddit posts. Furthermore, we gauge the positivity of generated film reviews by assessing them using the \textit{distilbert-imdb} model \cite{DistilBERT}.

\begin{table}[h]
\caption{Various tasks, input and output types, and the metrics used in the TIL settings.}
\resizebox{\linewidth}{!}{
\begin{tabular}{llllll}
\toprule
\textbf{Dataset}      & \textbf{Task}                                                                      & \textbf{Input}                                                 & \textbf{Output}                                                                          & \textbf{\begin{tabular}[c]{@{}l@{}}Preference \\ Metric\end{tabular}}          & \textbf{\begin{tabular}[c]{@{}l@{}}Naturalness \\ Metric(s)\end{tabular}}             \\ \midrule
\textbf{IMDB\cite{IMDB}}         & Text Continuation                                                                  & \begin{tabular}[c]{@{}l@{}}Partial Movie\\ Review\end{tabular} & \begin{tabular}[c]{@{}l@{}}A positive completion\\ of the movie review.\end{tabular}     & \begin{tabular}[c]{@{}l@{}}70M sentiment \\ classifier \textbf{DistilBERT}\end{tabular} & \multirow{3}{*}{\begin{tabular}[c]{@{}l@{}}RougeL, \\ BLEU-4, \\ METEOR\end{tabular}} \\ \cline{1-5}
\textbf{HH-RLHF\cite{HH_RLHF}}      & \begin{tabular}[c]{@{}l@{}}Helpful \& Harmless \\ Question Answering\end{tabular} & Question                                                       & \begin{tabular}[c]{@{}l@{}}A helpful and harmless \\ answer to the question\end{tabular} & \begin{tabular}[c]{@{}l@{}}2.7B \textbf{SteamSHP}-\\ flan-t5-xl model\end{tabular}      &                                                                                       \\ \cline{1-5}
\textbf{Reddit TL;DR\cite{summary_rlhf}} & Summarization                                                                      & Reddit POST                                                    & Summarized POST                                                                          & \begin{tabular}[c]{@{}l@{}}6.7B \textbf{GPT-J} reward \\ model by Carper-AI\end{tabular} &                                                                                       \\ \bottomrule
\end{tabular}
\label{tab:metrics4TIL}
}
\end{table}

\textbf{Domain Incremental Learning (DIL)} setting: The policy is required to continuously learn from three segments of the \textbf{SHP} dataset. The SHP dataset comprises 18 domains, which we divide into 3 parts  based on the highest observed performance decline. The details can be found in the Appendix Section \ref{sec:DILdesign}.
 We employ the \textit{SteamSHP-flan-t5-xl model} \cite{shp}, developed by Stanford, as the preference model (PM) for assessing responses to SHP prompts. The \textit{SteamSHP-flan-t5-xl model} is trained on the combination of the SHP (all domains) and the HH-RLHF human preference data.

\textbf{Evaluation Metric for Continual Learning}

\textbf{Overall performance} is typically evaluated by \textit{average accuracy} (AA)\cite{Rwalk,gem} and \textit{average incremental accuracy} (AIA) \cite{douillard2020podnet,hou2019learning}. 
In our evaluation setting, \textit{the accuracy is replaced by the Preference Metric} (0-1). 
Let ${a}_{k,j} \in [0,1]$ denote the Preference Score evaluated on the test set of the $j$-th task after incremental learning of the $k$-th task ($j \leq k$). The two metrics at the $k$-th task are then defined as
\begin{equation} {{\rm{AA}}_{k}} = \frac{1}{k} \sum_{j=1}^{k} {a}_{k,j}, 
\end{equation}
\begin{equation} 
{{\rm{AIA}}_{k}} = \frac{1}{k} \sum_{i=1}^{k} {\rm{AA}}_{i}, 
\end{equation}
where AA represents the overall performance at the current moment and AIA further reflects the historical variation.

\textbf{Memory stability} can be evaluated by \textit{forgetting measure} (FM)\cite{Rwalk} and \textit{backward transfer} (BWT) \cite{gem}. As for the former, the forgetting of a task is calculated by the difference between its maximum performance obtained in the past and its current performance:
\begin{equation}
f_{j,k} = \max_{i \in \{1,...,k-1\}} ({a}_{i,j} - {a}_{k,j}), \forall j<k.
\end{equation}
FM at the $k$-th task is the average forgetting of all old tasks:
\begin{equation}
    {{\rm{FM}}_{k}} = \frac{1}{k-1} \sum_{j=1}^{k-1} f_{j,k}.
\end{equation}
As for the latter, BWT evaluates the average influence of learning the $k$-th task on all old tasks: 
\begin{equation}
    {\rm{BWT}}_k = \frac{1}{k-1} \sum_{j=1}^{k-1} ({a}_{k,j} - {a}_{j,j}),
\end{equation}
where the forgetting is usually reflected as a negative BWT.

\subsection{Baselines}
\label{sec:baseline}

\textbf{Supervise fine-tuning (SFT)} directly learns the human-labeled summary through the cross-entropy loss. We combine SFT with classic continual learning methods.


\begin{itemize}
    \item \textbf{SFT-Online L2Reg} penalizes the updating of model parameters through an L2 loss $L_2^t(\theta) = \sum_{i} ({\theta}^i_{t} - {\theta}^i_{t-1})^2$.  This regularization term mitigates the forgetting issue by applying a penalty for every parameter change.
    \item \textbf{SFT-EWC} \cite{EWC} uses fisher information to measure the parameter importance to old tasks, then slows down the update of the important parameters by L2 regularization.
    \item \textbf{SFT-MAS} \cite{MAS} computes the importance of the parameters of a neural network in an unsupervised and online manner to restrict the updating of parameters in the next task.
    \item \textbf{SFT-AGM} \cite{AGM} is an improved version of GEM \cite{gem}, which enjoys better performance than GEM, while being almost as computationally and memory efficient as EWC and other  regularization-based methods. 
    \item \textbf{SFT-LwF} \cite{LWF} is a knowledge-distillation-based method, which computes a smoothed version of the current responses for the new examples at the beginning of each task, minimizing their drift during training.
    \item \textbf{SFT-TFCL} \cite{TFCL} proposes to timely update the importance weights of the parameter regularization by detecting plateaus in the loss surface.
    \item \textbf{SFT-DER++}\cite{NEURIPS2020_b704ea2c} addresses the General Continual Learning (GCL) problem by mixing rehearsal with knowledge distillation and regularization, in which the logits and ground truth labels of part of old data are saved into the memory buffer for replaying.
\end{itemize}






Recent alignment methods are not able to continually learn human preference, we improve those methods with continual tricks.

\textbf{Ranking-based Approachs}\cite{DPO,PRO,P3O,rrhf,raft} rank human preferences over a set of responses and directly incorporate the ranking information into the LLMs fine-tuning stage. 
\begin{itemize}
    \item \textbf{DPO}$^C$\cite{DPO} is an offline approach that can directly align LM with human preference data, drawing from the closed-form solution of the Contextual Bandit with the KL control problem. 
    \item \textbf{PRO}$^C$\cite{PRO} learns preference ranking data by initiating with the first preferred response, deems subsequent responses as negatives, and then dismisses the current response in favor of the next. 
    \item \textbf{RRHF}$^C$\cite{rrhf} aligns with human preference by a list rank loss and finds that the SFT training objective is more effective and efficient than KL-divergence in preventing LLMs from over-fitting. 
\end{itemize}

\textbf{Language-based Approach}  directly uses natural language to inject human preference via SFT.
\begin{itemize}
\item \textbf{CoH}$^C$\cite{CoH} directly incorporates human preference as a pair of parallel responses discriminated as low-quality or
high-quality using natural language prefixes. CoH only applies the fine-tuning loss to the actual model outputs, rather than the human feedback sequence and the instructions. During inference, CoH directly puts position feedback (e.g., good) after the input instructions to encourage the LLMs to produce high-quality outputs.
\end{itemize}

\subsection{Results of Continual Learning from Human Preferences}
\begin{figure}[H]
    \centering
    \subfigure[Performance on HH-RLHF]{
    \includegraphics[width=0.48\linewidth]{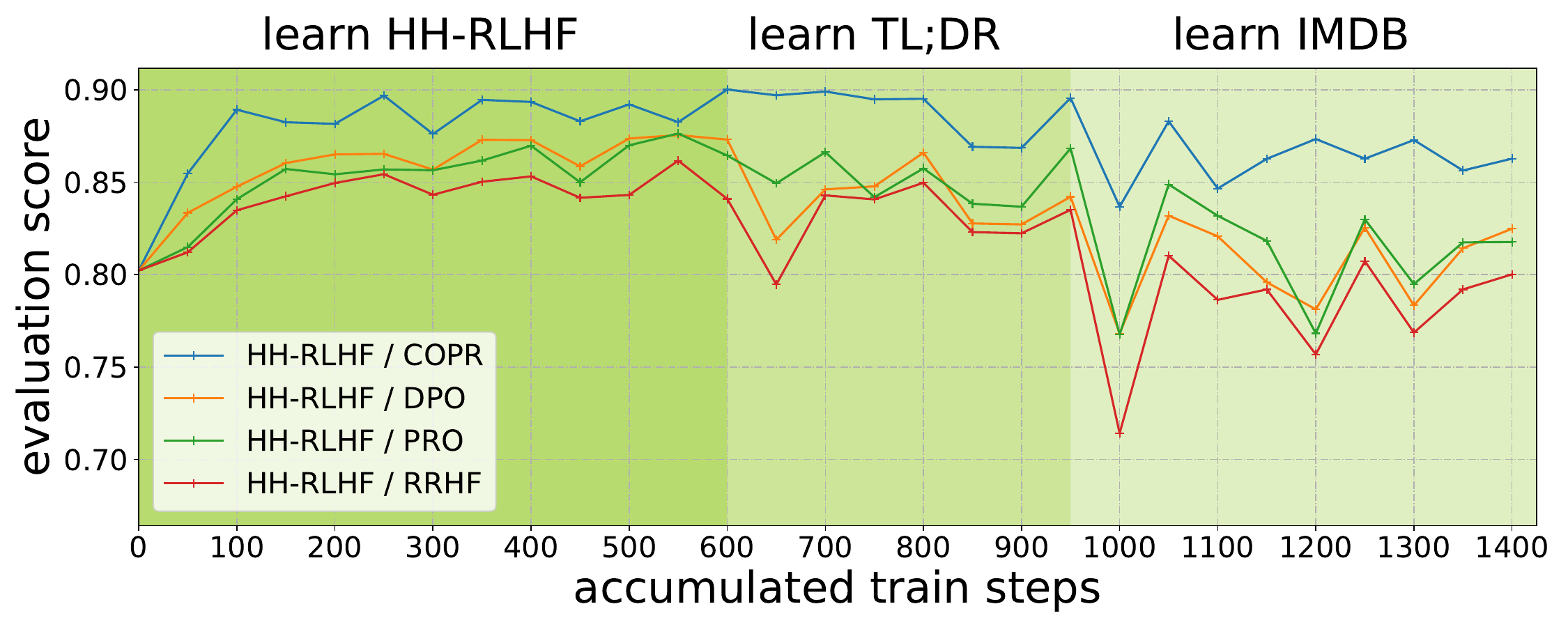}}
    \subfigure[Performance on Reddit TL;DR]{
    \includegraphics[width=0.48\linewidth]{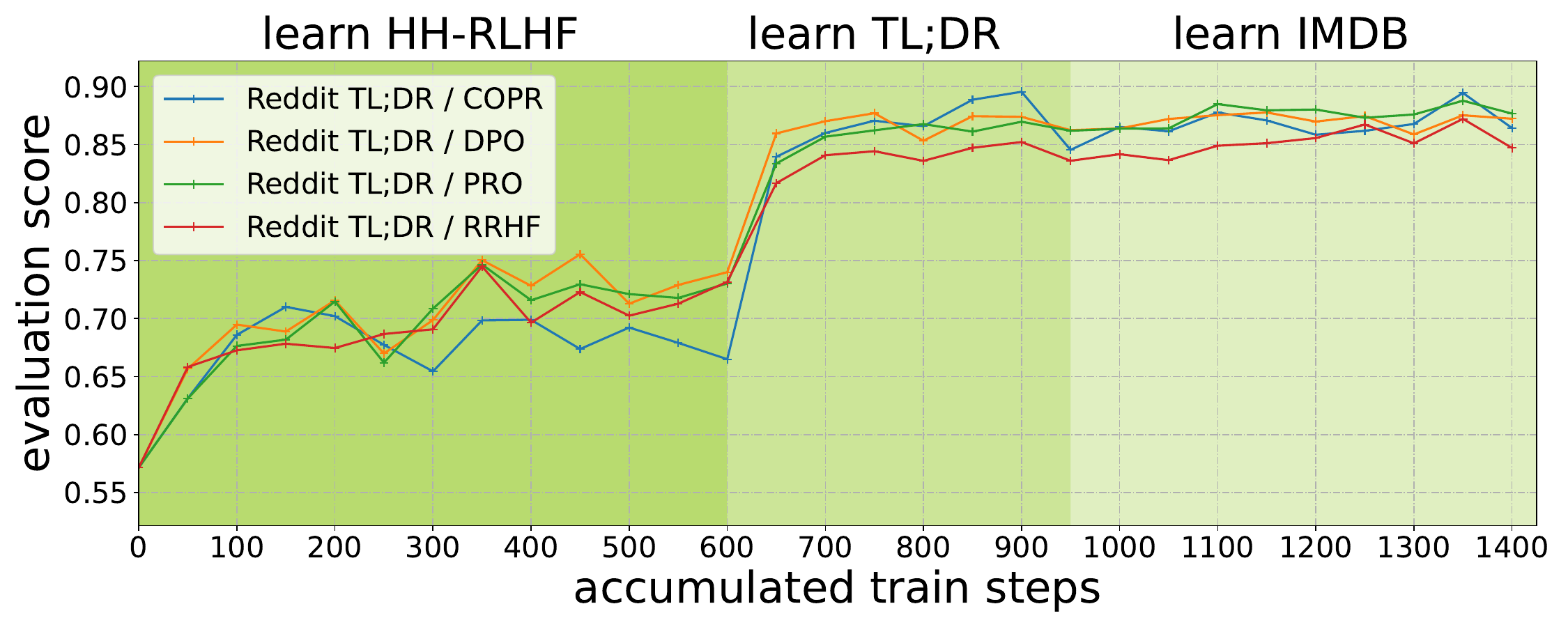}}
    \caption{Evaluation curves of TIL setting. Every 50 training steps, we evaluate the model on three tasks.}
    \label{fig:TIL-eval}
\end{figure}

Table \ref{tab:mainEXPs} shows the test results of continual learning from human preferences on the TIL setting.
Figure \ref{fig:TIL-eval} shows the valid results of continual learning from human preferences on TIL setting.

\begin{table}[h]
\centering
\caption{The performances at last task in the TIL scenario. \textbf{\textit{Italic}} indicates methods based on Supervised Fine-Tuning that did not use pair data and only utilized prompt and chosen data. For fair comparison, all baselines utilize 1\% historical replay data. COPR$^L$ and COPR$^G$ denote the linear and gaussian reward respectively. Due to the original CoH, DPO and PRO methods are not supported for continual learning, we introduce the CoH$^C$, DPO$^C$ and PRO$^C$ by adding 1\% historical replay samples into the training data. All of the experiments are based on the Llama-7b-hf. }
\resizebox{\linewidth}{!}{
\begin{tabular}{l|ccccccc}
\toprule
\multicolumn{1}{c}{\multirow{2}{*}{\textbf{Method / Taxonomy}}}              & \textbf{HH-RLHF}      & \textbf{Reddit TL;DR} & \textbf{IMDB}            & \multicolumn{2}{c}{\textbf{Overall performance}} & \multicolumn{2}{c}{\textbf{Memory stability}} \\
\multicolumn{1}{c}{}                                                         & \textbf{SteamSHP (↑)} & \textbf{Gpt-J (↑)}    & \textbf{DistillBert (↑)} & \textbf{AA (↑)}        & \textbf{AIA (↑)}        & \textbf{BWT (↑)}       & \textbf{FM (↓)}            \\ \midrule
\textit{\textbf{L2-Reg}} / function regularization                           & 0.797                 & 0.812                 & 0.812                    & 0.807                  & 0.807                   & -0.028                 & 0.031                \\
\textit{\textbf{EWC}}\cite{EWC} / weight regularization                     & 0.803                 & 0.821                 & 0.826                    & 0.817                  & 0.821                   & -0.026                 & 0.026                \\
\textit{\textbf{MAS}}\cite{MAS} / weight regularization                     & 0.807                 & 0.819                 & 0.812                    & 0.813                  & 0.819                   & -0.027                 & 0.027                \\
\textit{\textbf{AGM}}\cite{AGM} / gradient projection                     & 0.812                 & 0.827                 & 0.832                    & 0.824                  & 0.829                   & -0.022                 & 0.024                \\
\textit{\textbf{LwF}}\cite{LWF}  / function regularization                  &                       0.813&                       0.841&                          0.833&                        0.829 &                         0.820 &                        -0.024&                      0.024 \\
\textit{\textbf{TFCL}}\cite{TFCL} /  weight regularization             & 0.813 & 0.832 & 0.829 & 0.825 & 0.821 & -0.021 & 0.021 \\  
\textit{\textbf{DER++}}\cite{NEURIPS2020_b704ea2c} / experience replay     & 0.815 & 0.837 & 0.841 & 0.831 & 0.836 & -0.017 & 0.019 \\ 
\midrule
\textbf{CoH$^C$}\cite{CoH} / experience replay     & 0.807 & 0.883 & 0.825 & 0.838 & 0.854 & -0.027 & 0.027 \\
\textbf{PRO$^C$}\cite{PRO} / experience replay    & 0.795 & 0.876 & 0.854 & 0.842 & 0.857 & -0.038 & 0.038               \\
\textbf{RRHF$^C$}\cite{rrhf} / experience replay                          & 0.807 & 0.867 & 0.861 & 0.845 & 0.848 & -0.020 & 0.020                \\
\textbf{DPO$^C$}\cite{DPO} / experience replay& 0.825                 & 0.875                 & 0.844                    & 0.848                  & 0.862          & -0.026                 & 0.026                \\
\textbf{COPR$^L$ / function regularization} & 0.836          & 0.877          & \textbf{0.876} & \textbf{0.863} & 0.851          & \textbf{0.006} & \textbf{-0.006} \\
\textbf{COPR$^G$ / function regularization} & \textbf{0.856} & \textbf{0.895} & 0.815          & 0.855          & \textbf{0.878} & -0.021         & 0.021  \\
\bottomrule
\end{tabular}
}
\label{tab:mainEXPs}
\end{table}

\subsection{Learning unlabeled responses}
\label{sec:unlabel}

In this section, we discuss how the COPR policy makes use of unlabeled prompts. Standard RLHF methods can leverage additional unlabeled prompts by labeling LM generations with the learned reward model. Inspired by the learned reward model, we introduce the reward value head, namely 
the \textit{linear reward score} mode, to learn the human preference scoring on the labeled data at the meanwhile of optimal policy fitting process. We find that the two objectives are not in conflict.

\begin{table}[H]
\caption{Performance on DIL setting in which part of data is unlabeled.}
\resizebox{\linewidth}{!}{
\begin{tabular}{c|ccccccc}
\toprule
                                  & \textbf{Domains 1-6}&\textbf{ Domains 7-12}& \textbf{Domains 13-18}& \multicolumn{2}{c}{\textbf{Overall performance}} & \multicolumn{2}{c}{\textbf{Memory stability}}                            \\
\multirow{-2}{*}{\textbf{Method}} & \textbf{SteamSHP (↑)} & \textbf{SteamSHP (↑)} & \textbf{SteamSHP (↑)} & \textbf{AA (↑)}        & \textbf{AIA (↑)}        & \textbf{BWT (↑)}               & \textbf{FM (↓)}                         \\ \midrule
\textbf{COPR$^L$ / label}& 0.881                 & 0.901                 & 0.909                    & 0.897                  & 0.873                   & 0.027                  & -0.015               \\
\textbf{COPR$^L$ / task-2 unlabel}& \textbf{0.881}        & \textbf{0.912}        & \textbf{0.913}           & \textbf{0.902}         & \textbf{0.874}          & \textbf{0.033}         & \textbf{-0.020}               \\
\textbf{COPR$^L$ / task-1,2 unlabel}& 0.873                 & 0.883                 & 0.904                    & 0.887                  & 0.870                   & 0.015                  & 0.000        \\
\bottomrule
\end{tabular}}
\label{tab:unlabel}
\end{table}

After learning about the labeled data, we utilize the LM with the value head as explicit RM. We suppose that for each prompt $x$, there are multiple responses $\mathbf{{Y}}^x = \{y_1^x, y_2^x,..., y_{J_x}^x \}$ without knowing the human preference order. The responses can be generated by the trained policy or collected by other models or humans, like the PPO method. Then we utilize the explicit reward model to sort responses in set $\mathbf{{Y}}^x$. We conduct an experiment on the SHP dataset under the DIL setting. The LM is trained in the first 6 domains to learn the ability to score and to align with human preference. Then the LM with a value head is treated as an explicit RM to score the the unlabeled pairwise data. Here, we utilize the same prompts and responses in the SHP data, but the preference order is sorted by the explicit RM.

 
 In Table \ref{tab:unlabel}, we compare the learning results on unlabeled prompts and on the original SHP data. An observation from the learning results is that learning on unlabeled task-2 has better performance than learning on all labeled tasks.

\section{Related Work}

\subsection{Taxonomy of representative continual learning methods}
As illustrated in Figure \ref{fig: taxonomy}, within the realm of continual learning, several noteworthy methodologies emerge, encompassing the\textit{ regularization-based approach, replay-based techniques, optimization-based strategies, representation-based methodologies}, and \textit{architecture-based innovations }\cite{SurveyCL2023}.

\textbf{The Regularization-Based Approach} orchestrates the introduction of explicit regularization terms, thereby striving to strike a harmonious balance between the acquisition of new skills and the retention of past knowledge. 
Weight regularization \cite{EWC,MAS,Rwalk}is concerned with tempering the fluctuation of network parameters, often achieved by integrating a quadratic penalty term into the loss function. 
In contrast, \textit{function regularization}\cite{LWF,castro2018end} turns its focus to the intermediate or final output of the prediction function.

\textbf{The Replay-Based Approach} presents a multifaceted strategy encompassing three distinct sub-directions. The initial approach, \textit{experience replay}\cite{experience_replay}, revolves around the retention of a limited cache of prior training samples within a compact memory buffer. The second approach, \textit{generative replay or pseudo-rehearsal} \cite{Lamol}, entails the training of an additional generative model tasked with reproducing synthetic data samples. In contrast, \textit{feature replay}\cite{liu2020generative} diverges from its counterparts by maintaining feature-level distributions, bypassing the replication of entire data samples. 

\textbf{The Optimization-Based Approach} navigates the terrain of continual learning through explicit design and manipulation of optimization programs. This includes techniques such as \textit{gradient projection}\cite{gem}, and \textit{meta-learning}\cite{NEURIPS2019_f4dd765c}. Gradient projection serves as a mechanism for constraining parameter updates, ensuring alignment with the trajectory of experience replay. 
Meta-learning, often referred to as "learning-to-learn" within the context of continual learning, endeavors to cultivate a data-driven inductive bias adaptable to various scenarios, thus eliminating the need for manual design.

\textbf{The Representation-Based Approach} leverages the strengths of self-supervised learning (SSL)\cite{gallardo2021self} and large-scale pre-training\cite{mehta2022an} to enhance the quality of representations at both the initialization and continual learning stages. This category encompasses \textit{self-supervised learning} (primarily employing contrastive loss) for continual learning, \textit{pre-training for downstream} continual learning, and \textit{continual pre-training} (CPT)\cite{han-etal-2021-econet} or continual meta-training\cite{mbpa}.

\textbf{The Architecture-Based Approach} addresses inter-task interference by fashioning task-specific parameters. This approach can be dissected into three distinct paradigms: \textit{parameter allocation}\cite{serra2018overcoming}, where dedicated parameter subspaces are allocated to each task, either maintaining a fixed or dynamic network architecture; \textit{model decomposition}\cite{ebrahimi2020adversarial}, which explicitly segregates a model into task-sharing and task-specific components, with the latter typically being expandable; and \textit{modular networks}\cite{rusu2016progressive}, which harness parallel sub-networks or sub-modules to facilitate differentiated learning of incremental tasks, devoid of pre-defined task-sharing or task-specific components.

\subsection{Learning from Human Preferences}

Learning from human preferences has been studied in the game field \cite{RLHF_early1,RLHF_early2,RLHF_early3,RLHF_early4} and has recently been introduced into the NLP domain. Previous work \cite{summary_rlhf} utilizes the PPO algorithm to fine-tune a language model (LM) for summarization and demonstrates that RLHF can improve the LM's generalization ability, which serves as the technology prototype for InstructGPT \cite{InstructGPT} and ChatGPT \footnote{A dialogue product of OpenAI: \url{https://openai.com/blog/chatgpt}}. Learning LMs from feedback can be divided into two categories: human or AI feedback. Recent works such as HH-RLHF \cite{HH_RLHF} and InstructGPT \cite{InstructGPT} collect human preferences to train a reward model and learn a policy through it. ILF \cite{ILF} proposes to learn from natural language feedback, which provides more information per human evaluation. Since human annotation can be expensive, learning from AI feedback (RLAIF) \cite{RLAIF,RedTeam_deepmind,RedTeam_Anthropic} is proposed, but current methods are only effective for reducing harmless outputs, while helpful outputs still require human feedback. 
Furthermore, recent works study the empirical challenges of using RL for LM-based generation. NLPO \cite{NLPO} proposes to improve training stability and exhibits better performance than PPO for NLP tasks by masking invalid actions. \cite{rm_scaling} investigates scaling laws for reward model overoptimization when learning from feedback.

Preference-based Reinforcement Learning (PbRL) relies on binary preferences generated by an unknown 'scoring' function, as opposed to conventional reward-based learning \cite{busa2014preference,ruiz2023dueling}. While various PbRL algorithms are available, some of which can recycle off-policy preference data, they typically begin by explicitly estimating the latent scoring function (i.e., the reward model) and subsequently optimizing it \cite{jain2013learning,busa2014preference,christiano2017deep,sadigh2017active,kupcsik2018learning}.
Due to the complexity of the RLHF pipeline and the high memory demands and suboptimal utilization of computational resources by the PPO algorithm, recent works \cite{DPO,raft,rrhf} begin exploring \textit{non-reinforcement learning} methods to directly learn human preferences.

\section{Conclusion}
While Reinforcement Learning from Human Feedback is a widely-used technique to enhance pre-trained Language Models (LM) for better alignment with human preferences, the need for full retraining when introducing new queries or feedback presents significant challenges. This retraining process is impractical in many real-world scenarios due to time, computational, and privacy concerns. Our proposed method, Continual Optimal Policy Regularization, overcomes these limitations, offering a single-phase, reinforcement learning-free approach that effectively aligns with human preferences across various tasks and domains, making it a promising advancement in preference learning.  The experiments indicate that COPR can outperform existing continual learning methods in both task and domain continual learning scenarios.

\bibliographystyle{unsrt}  
\bibliography{references}  

\appendix
\section{Theory analysis}
\subsection{The optimization of pairwise ranking loss}
\label{sec:pairwise}
The gradients of $\mathcal{L}_{ranking}=-\log(\sigma(r_\theta(x,y_w)-r_\theta(x,y_l)))=-\log(\sigma(r_w-r_l))$ according to $r_w$ and $r_l$ respectively are:

\begin{equation}
    \frac{\partial\mathcal{L}_{ranking}}{\partial{r_w}} = -(1-\sigma(r_w-r_l))
\label{eq:gradW}
\end{equation}

\begin{equation}
    \frac{\partial\mathcal{L}_{ranking}}{\partial{r_l}} = 1-\sigma(r_w-r_l)
\label{eq:gradL}
\end{equation}
Considering that the partially-ordered set $\mathcal{Y}^x = \{y_1^x \prec y_2^x \prec...\prec y_{J_x}^x \}$, according to Eq \ref{eq:gradW}  and Eq \ref{eq:gradL}, the accumulation of gradient according to $r_j$ is 
\begin{equation}
G_j = \Sigma_{k=1}^{j-1} -(1-\sigma(r_j-r_k)) + \Sigma_{k=j+1}^{J_x} (1-\sigma(r_{k}-r_j))
\end{equation}
where  $r_k$ $(k=1,2,...,J_x)$  denotes the reward score of response $y_k^x$.
We suppose that the initial reward $r_j$ is close to zero. 
In the early stages of training, the reward value $r_j$ is approximated to $0 -\eta\cdot{G_i} \approx (j-1)\cdot0.5\eta - (J_x-j)\cdot0.5\eta = \eta \cdot j - 0.5\eta (J_x + 1)$ which exhibits a linear relationship with the degree of human preference $j$.

\section{Datasets}
\label{sec:dataset}

\textbf{Stanford Human Preferences (SHP) Dataset}\cite{shp}: SHP is a dataset of 385K collective human preferences over responses to questions/instructions in 18 different subject areas, from cooking to legal advice. The preferences are meant to reflect the helpfulness of one response over another and are intended to be used for training RLHF reward models and NLG evaluation models (e.g., SteamSHP).


\textbf{Helpful and Harmless (HH) \cite{HH_RLHF}}: The HH-RLHF dataset is collected by two separate datasets using slightly different versions of the user interface. The helpfulness dataset is collected by asking crowdworkers to have open-ended conversations with models, asking for help, and advice, or for the model to accomplish a task, and to choose the more helpful model response. The harmlessness or red-teaming dataset is collected by asking crowd workers to attempt to elicit harmful responses from our models and to choose the more harmful response offered by the models. 

\textbf{Reddit TL;DR}:  For each Reddit post in the Reddit TL;DR  \cite{TLDR} dataset, multiple summaries are  generated using various models. These models include pre-trained ones used as zero-shot summary generators, as well as supervised fine-tuned models (12B, 6B, and 1.3B) specifically trained on the Reddit TL;DR dataset. Additionally, the human-written TL;DR (reference) is considered as a sample for comparison.

\textbf{IMDB}: We consider the IMDB dataset for the task of generating text with positive sentiment. The IMDB text continuation task aims to positively complete the movie review when given a partial review as a prompt. In this task, a trained  sentiment classifier DistilBERT\cite{DistilBERT}  is provided as a reward function to train the RL agents and  evaluate their task performance. The naturalness of the trained model is evaluated with a perplexity  score. The dataset consists of 25k training, 5k validation, and 5k test examples of movie review text with sentiment labels of positive and negative. The input to the model is a partial movie review text (up to 64 tokens) that needs to be completed (generating 48 tokens) by the model with a positive sentiment while retaining fluency. For RL methods, we use a sentiment classifier that is trained on pairs of text and labels as a reward model which provides sentiment scores indicating how positive a given piece of text is.

\section{Task design for DIL}
\label{sec:DILdesign}

\begin{figure}
    \centering
    \includegraphics[width=\linewidth]{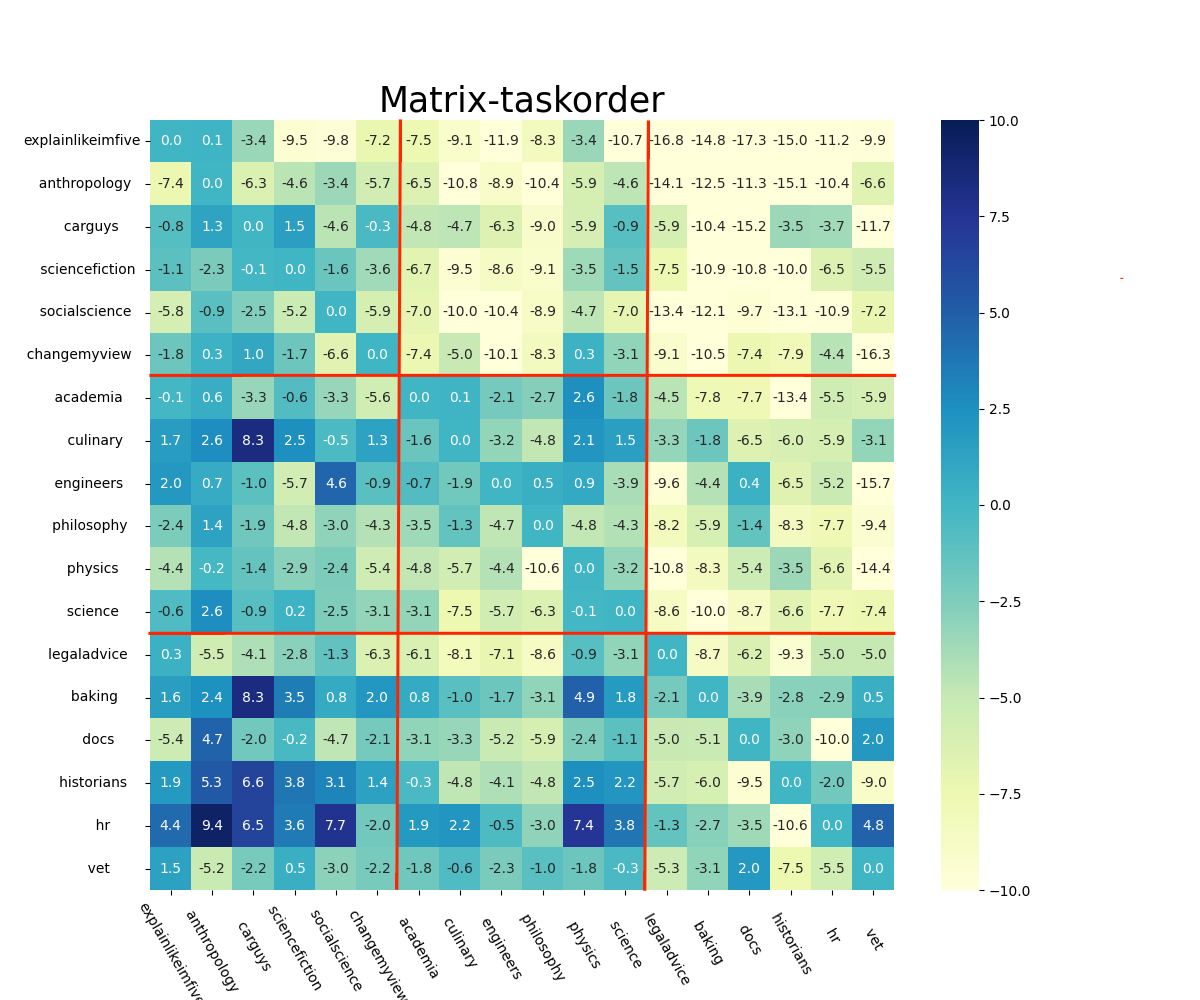}
    \caption{Cross-domain evaluation rewards the model. Rows represent the training set domain-i for the reward model, columns represent the testing set domain-j for evaluating the model, and the values indicate the accuracy degradation data of the model in domain-j testing compared to domain-i testing. }
    \label{fig:DILMatrix}
\end{figure}

To create more challenging DIL tasks, we individually trained 18 reward models (based on Llama-7b) on 18 domains of SHP data and evaluated each reward model on the test set of the 18 domains, resulting in an accuracy difference matrix of size 18x18 as shown in Figure \ref{fig:DILMatrix}. 
In this matrix, the row coordinates represent the training domains, and the column coordinates represent the evaluation domains. The elements in the matrix represent the relative decrease in performance on the evaluation domain compared to the training domain (both evaluated on test sets from various domains). 
Based on this accuracy difference matrix, we divided the 18 domains into 3 groups (each has 6 domains). This division ensures that there will be a significant performance decrease, i.e., larger out-of-group generalization error, when evaluated on domains from different groups.

For example, the RM trained on the first gourp (explainlikeimfive, anthropology, ..., changemyview) has large performance decrease on the third group (legaladivce, baking, ..., vet).

\end{document}